\documentclass[journal,twoside,web]{ieeecolor}
\usepackage{tmi}
\usepackage{cite}
\usepackage{amsmath,amssymb,amsfonts}
\usepackage{algorithmic}
\usepackage{graphicx}
\usepackage{textcomp}

\let\labelindent\relax
\usepackage{enumitem}
\usepackage{amsmath, amssymb,bm}
\usepackage{booktabs}
\usepackage{graphicx,epstopdf,subfigure}
\usepackage{multirow}
\usepackage{tabularx,threeparttable}
\usepackage{array}
\usepackage{comment}
\usepackage[hidelinks]{hyperref}
\usepackage{mathtools}
\usepackage{bbm}
\usepackage[ruled]{algorithm2e}
\usepackage{dsfont}
\usepackage{arydshln}

\makeatletter
\DeclareRobustCommand\onedot{\futurelet\@let@token\@onedot}
\def\@onedot{\ifx\@let@token.\else.\null\fi\xspace}

\def\eg{\emph{e.g}\onedot} 
\def\ie{\emph{i.e}\onedot}

\makeatother

\def\BibTeX{{\rm B\kern-.05em{\sc i\kern-.025em b}\kern-.08em
    T\kern-.1667em\lower.7ex\hbox{E}\kern-.125emX}}
\begin{document}
\title{Exploring Intra- and Inter-Video Relation for Surgical Semantic Scene Segmentation}
\author{Yueming Jin,~\IEEEmembership{Member,~IEEE}, Yang Yu, Cheng Chen,~\IEEEmembership{Member,~IEEE}, Zixu Zhao,~\IEEEmembership{Student Member,~IEEE}, \\ Pheng-Ann Heng,~\IEEEmembership{Senior Member,~IEEE}, Danail Stoyanov,~\IEEEmembership{Senior Member,~IEEE}
\thanks{Manuscript received March 16, 2022; revised May 7, 2022; accepted May 10, 2022.
This research was funded in whole, or in part, by the Wellcome/EPSRC Centre for Interventional and Surgical Sciences (WEISS) [203145/Z/16/Z]; the Engineering and Physical Sciences Research Council (EPSRC) [EP/P027938/1, EP/R004080/1, EP/P012841/1]; the Royal Academy of Engineering Chair in Emerging Technologies Scheme, and Horizon 2020 FET Open (863146); Key-Area Research and Development Program of Guangdong Province, China under Grant 2020B010165004; Hong Kong RGC TRS Project No. T42-409/18-R. For the purpose of open access, the author has applied a CC BY public copyright licence to any author accepted manuscript version arising from this submission.}
\thanks{Yueming Jin and Danail Stoyanov are with the Wellcome/EPSRC Centre for Interventional and Surgical Sciences (WEISS) and the Department of Computer Science, University College London. (e-mail: \{yueming.jin, danail.stoyanov\}@ucl.ac.uk).}
\thanks{Yang Yu, Cheng Chen, Zixu Zhao and Pheng-Ann Heng are with the Department of Computer Science and Engineering, The Chinese University of Hong Kong, HKSAR, China. Pheng-Ann Heng is also with Guangdong Provincial Key Laboratory of Computer Vision and Virtual Reality Technology, Shenzhen Institutes of Advanced Technology, Chinese Academy of Sciences, Shenzhen, China. (e-mail: \{yangyu, cchen, zxzhao, pheng\}@cse.cuhk.edu.hk).}
}

\vspace{-10mm}
\maketitle

\begin{abstract}
Automatic surgical scene segmentation is fundamental for facilitating cognitive intelligence in the modern operating theatre. 
Previous works rely on conventional aggregation modules (\eg, dilated convolution, convolutional LSTM), which only make use of the local context.
In this paper, we propose a novel framework \emph{STswinCL} that explores the complementary intra- and inter-video relations to boost segmentation performance, by progressively capturing the global context.
We firstly develop a hierarchy Transformer to capture intra-video relation that includes richer spatial and temporal cues from neighbor pixels and previous frames. A joint space-time window shift scheme is proposed to efficiently aggregate these two cues into each pixel embedding.
Then, we explore inter-video relation via  pixel-to-pixel contrastive learning, which well structures the global embedding space. 
A multi-source contrast training objective is developed to group the pixel embeddings across videos with the ground-truth guidance, which is crucial for learning the global property of the whole data.
We extensively validate our approach on two public surgical video benchmarks, including EndoVis18 Challenge and CaDIS dataset. Experimental results demonstrate the promising performance of our method, which consistently exceeds previous state-of-the-art approaches.
\textcolor{black}{Code is available at \url{https://github.com/YuemingJin/STswinCL}.}

\end{abstract}


\section{Introduction}

\IEEEPARstart{C}{omputer} assisted interventions (CAI) are revolutionizing surgical procedures to achieve enhanced patient safety with improved operative quality, reduced adverse events and shorter recovery periods~\cite{maier2017surgical}. 
Semantic segmentation of the entire surgical scene in the field of view of the surgical camera is an essential prerequisite for modern CAI systems. Semantic labels can facilitate cognitive assistance, by providing pixel-wise context awareness of tissues and instruments, which is fundamentally required for supporting several downstream tasks, such as surgical decision making~\cite{loftus2020artificial,maier2022surgical}, surgical navigation~\cite{allan20202018} and skill assessment~\cite{curtis2020association,Liu_2021_CVPR}.
Precisely identifying instruments and their location is also a central CAI theme with work on tool pose estimation~\cite{hein2021towards}, tool tracking and control~\cite{du2019patch} and surgical task automation~\cite{nagy2019dvrk}.  
However, precisely parsing the entire scene from surgical video is highly challenging because complicated surgical scenes lead to limited inter-class variance (\eg different instruments) and high intra-class variance (\eg covered kidney). Class imbalance also exists in surgical scenes, in which tiny objects (\eg thread) and rarely used instruments are difficult to identify. Motion blur, lighting changes, occlusions from smoke and blood, further increase the challenges that segmentation models need to overcome.

\begin{figure}[t]
	\centering
	\includegraphics[width=0.5\textwidth]{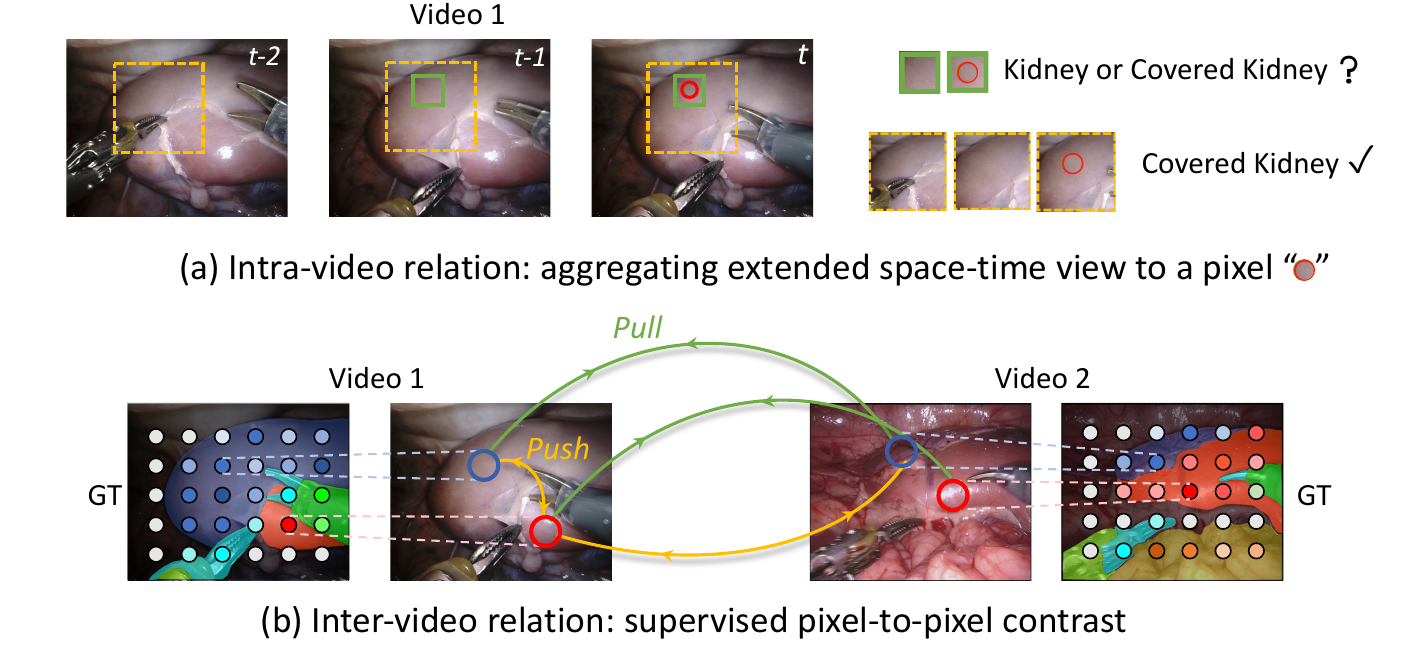}
	\vspace{-5mm}
	\caption{Two core factors for accurate surgical scene segmentation.
	(a) High discrimination of each pixel embedding via \emph{intra-video relation modelling}: aggregating richer space-time context to each pixel.
	(b) Well-structured semantic embedding space via \emph{inter-video relation modelling}: computing pixel-to-pixel contrast by pulling pixels from the same class closer while repelling pixels of different classes.
    We consider both relations in model learning to boost segmentation.
	}
	\label{fig:intro}
	\vspace{-6mm}
\end{figure}

We identify two main limitations in current state-of-the-art literature to tackle this task.
Firstly, most recent works rely on traditional feature aggregation modules, such as atrous convolution, whose encoded visual context is still relatively local for segmenting surgical scenes.
Meanwhile they solely model the spatial information while ignore the sequential dynamics in surgical videos~\cite{allan20202018,grammatikopoulou2021cadis,ren2020task}. The
conventional temporal modelling module of convolutional LSTM is just employed in a recent study~\cite{wang2021noisy}, which is also limited by the local reception field, otherwise suffers from model training difficulty.
The perceived context in these works is insufficient to achieve accurate segmentation for complicated surgical scenes.
See example in Fig.~\ref{fig:intro}(a), classifying pixel (red circle) is difficult by watching local space-time context (green boxes) due to limited inter-class variance between kidney and covered kidney.
Extending space-time view (yellow boxes) to see the folded fascia helps accurately categorize the pixel to covered kidney.
In this regard, it is of great importance to leverage \emph{\textbf{abundant intra-video context}} in both spatial and temporal dimensions to enhance feature representation capability for accurate segmentation.
Recently, Transformer-based model emerges to relate richer cues to each pixel by self-attention, which has demonstrated its efficacy on single image scene parsing task in computer vision field~\cite{zheng2021rethinking,liu2021swin}.
However, relating more view by self-attention shall lead to quadratic computational complexity increase with respect to the number of pixel embeddings~\cite{liu2021swin}.
Our scene segmentation of surgical video needs a dense relation from multiple frames, bringing a more immense set of pixels.
How to efficiently relate richer view is still an open problem that needs to be solved.

The other limitation of prior work is ignoring global context of the whole dataset, since they purely use information from the current video, and employ the de facto cross entropy loss to penalize each pixel embedding independently towards ground-truth.
The global semantic space is therefore not compact with ambiguous decision boundary, which cannot differentiate similar pixels from different classes or consistently categorize the dissimilar pixels from the same class. %
However, this challenge generally appears in surgical scenes, which are more severe when pixels come from different videos (\eg, see two blue circles in Fig.~\ref{fig:intro}(b), consistently segmenting covered kidney is difficult with dissimilar appearances).
Additionally, only performing pixel-to-groundtruth regularization in current literature incurs another inferiority of embedding space: the decision boundary is drastically altered by majority classes due to label bias in class-imbalanced surgical scene.
Therefore, it is highly demanded to explore \emph{\textbf{inter-video pixel-to-pixel relations}}, which can leverage the global semantic of pixels to better structure the embedding space. Recent advances in contrastive learning shares the similar insight of ``regularize embedding space by comparing sample pairs'': attracting positive pairs and repulsing negatives in the projected space.
Efficacy of well-shaped space has been demonstrated with successes in various tasks~\cite{chen2020simple,zhao2020contrastive}, which motivates us to explore the potential of contrastive learning for improving scene parsing in surgery scenario.

In this paper, we present a novel intra- and inter-video relation modelling for accurate surgical scene segmentation. Two types of relations are explored to boost segmentation by gradually capturing the global context of surgical videos.
Concretely, we first develop a novel Transformer-based model to explore intra-video relations, \ie, relating broad spatial-temporal view to each pixel, for strengthening its discrimination capability. To decrease computational cost, self-attention is only performed within a local space-time window, and depends on the joint space-time window shift scheme to efficiently expand the view boundary.
We then extend the relation modelling from the intra-video to inter-video, \ie, comparing the semantic similarities among labeled pixels from the whole dataset.
It is achieved by introducing a new pixel-to-pixel contrast objective, where the pair construction is guided by the ground-truth labels and selected from multiple sources across different videos.
By pulling the pixel pairs from the same class closer while repelling pixels of different classes, our contrast can capture the global property of the embedding space for better reflecting intrinsic structures of whole data, therefore enabling more accurate segmentations.
We extensively evaluate our method on two publicly available datasets, including a robotic surgery benchmark EndoVis18 and a cataract dataset CaDIS containing three sub-tasks, where our new approach outperforms existing state-of-the-art (SOTA). Our main contributions are:

\begin{enumerate}
\item
We propose a novel framework, \ie, \emph{STswinCL}, which jointly explores the intra- and inter-video pixel relations with two complementary training objectives.

\item
We develop a Transformer model, in which self-attention with joint \emph{\textbf{S}}pace-\emph{\textbf{T}}ime \emph{\textbf{s}}hifted \emph{\textbf{win}}dow (\emph{STswin}) is proposed to efficiently relate richer clues from intra-video to each pixel, for enhancing pixel category discrimination.

\item
We introduce a new structured comparison objective for the inter-video supervised \emph{\textbf{C}}ontrastive \emph{\textbf{L}}earning (\emph{CL}), which can explicitly enforce a global regularization to better structure the pixel semantic embedding space.

\item
Our method achieves performance gains in two open benchmark datasets improving on prior state-of-the-arts. 

\end{enumerate}

\section{Related Work}

Related literature on semantic segmentation in surgical video mainly focuses on surgical instrument segmentation, by exploring network architectures with holistically-nested networks~\cite{garcia2017toolnet}, attention mechanisms~\cite{ni2020pyramid}, multi-task learning with detection~\cite{sanchez2021scalable}, graph-based network~\cite{liu2021graph}, or taking advantage of additional cues including depth maps~\cite{mohammed2019streoscennet}, optical flow~\cite{jin2019incorporating}, motion flow~\cite{zhao2020learning}, \textcolor{black}{and synthesised images~\cite{colleoni2021robotic}.
The most related work is~\cite{wang2021efficient}, where the non-local mechanism and active memory module are designed to gather the global semantic correlation in long temporal range to boost instrument segmentation.}
Full scene segmentation is less explored with early methods using convolutional neural networks (CNNs) with promising results, using models like DeepLabV3+ to model the multi-scale information for robotic scene segmentation~\cite{allan20202018}, and HRNet to learn the high resolution representation for cataract scene parsing~\cite{grammatikopoulou2021cadis}.
Recently, Ren et al. decompose the single segmentation task into three subsequent tasks at different perceptual levels, and enable their interaction by task-task context ensemble~\cite{ren2020task}.
Pissas et al. propose the oversampling and adaptive sampling to tackle the class imbalance problem in surgical scene segmentation~\cite{pissas2021effective}.
Ni et al. propose a bilinear squeeze reasoning network to improve the discrimination capability of similar feature in the scene~\cite{ni2022space}.
Wang et al. use ConvLSTM on the video sequence, and add noise to one frame within sequence to spoil temporal coherence for improving model robustness~\cite{wang2021noisy}.
However, these methods mainly treat surgical video as static data, or simply leverage temporal cues by conventional modules, which easily encounters training difficulty.

\begin{figure*}[t]
	\centering
	\includegraphics[width=0.85\textwidth]{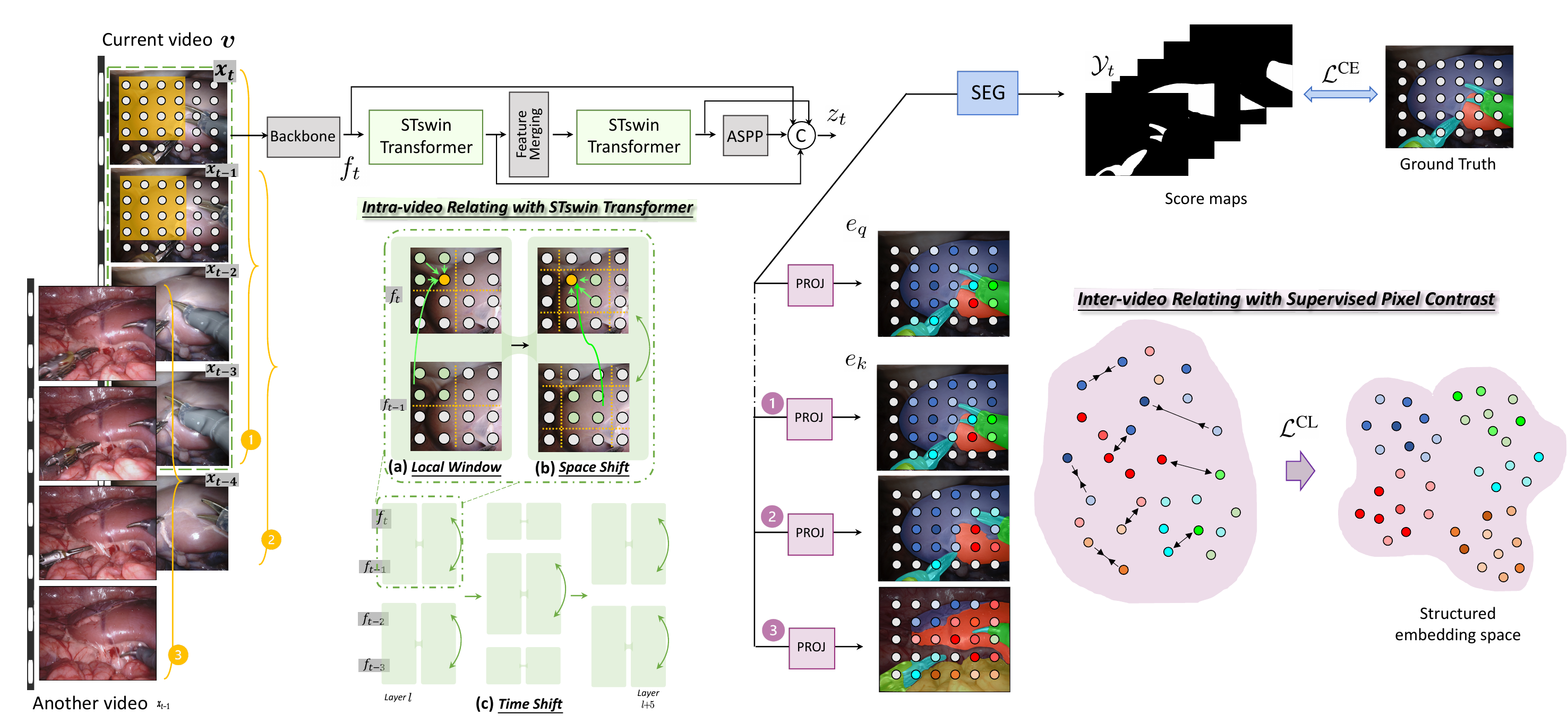}
	\vspace{-3mm}
	\caption{Schematic illustration of our intra and inter-video relation model for whole scene segmentation from surgical videos. 
	Specifically, given the current frame $x_t$, we establish two relations to boost segmentation from different aspects.
	(i) Forming a video clip $\{ x_{t-3}, \ldots, x_t \}$, our joint space-time shift Transformer fosters the differentiated capability of each pixel embedding from $f_t$ to $z_t$, by relating the rich \emph{intra-video information} from neighbor pixels and adjacent frames.
	(ii) Apart from de facto loss $\mathcal{L}^{\text{CE}}$, we compare the pixel-to-pixel relation to explicitly regularize the embedding space to a better structure via a new supervised contrastive learning $\mathcal{L}^{\text{CL}}$.
	Regarding pixel embeddings in current frame $e_q^i$ as query sample, \emph{inter-video contrast} is designed by collecting key samples $e_k^i$ from \textit{three} sources, which expands the embedding space to be shaped.} 
	\label{fig:overview}
	\vspace{-5mm}
\end{figure*}

Transformer~\cite{vaswani2017attention} is originally applied in the natural language processing field and recently has been applied to various vision problems~\cite{dosovitskiy2020image,carion2020end,meinhardt2021trackformer}.
The self-attention mechanism in it can consider the broad context to augment each token (\eg a pixel embedding in feature map which captures a local reception field) by learning the correlation between all the other token elements with it. To this end, Transformer can encode global dependencies beyond the local field in an image.
Existing works for natural scene segmentation still focus on exploring the relation within a single image~\cite{zheng2021rethinking,liu2021swin}. Recently, methods for action recognition~\cite{arnab2021vivit} and object tracking~\cite{meinhardt2021trackformer} further extend the range of encoded dependency to adjacent images. 
In surgical data science, Transformer has been used for tool detection~\cite{kondo2020lapformer} and workflow recognition~\cite{gao2021trans,czempiel2021opera,nwoye2022rendezvous}, while its efficacy for scene segmentation is not explored so far. One underlying reason may be the huge computational cost of the self-attention for this dense prediction task. A Transformer-based model~\cite{liu2021swin}, which limits self-attention to non-overlapping local windows while captures wider information by cross-window connection has been shown to achieve leading results with lightweight computation in natural semantic segmentation.
Although this work also only uses space cues in an image, the great efficiency brought by window connection inspires us to explore a joint space-time relation scheme for cost-effectively utilizing richer cues.

Contrastive learning has shown compelling capability to learn feature representations~\cite{hjelm2018learning,chen2020simple}. 
Similar to exemplar learning~\cite{dosovitskiy2014discriminative}, contrastive methods regularize model in a discriminative manner by attracting the similar (positive) pairs and repulsing the dissimilar (negative) pairs.
One major research direction of existing studies is how to construct positive and negative pairs, given it core role in representation learning.
For image data, standard positive pair selection is to perform perturbations to create multiple views of each image, while negative pair is generated from different images.
It has widely achieved promising performance on unsupervised representation learning of image-level tasks, such as image recognition and retrieval~\cite{chen2020simple,he2020momentum,grill2020bootstrap}, video action recognition~\cite{han2020self}.
Recently, pixel-level contrast emerges to tackle the dense prediction task.
Most works construct pairs based on the spatial distance, under assumption that neighbor pixels generally belong to the same class and form positive pair~\cite{xie2021propagate,chaitanya2020contrastive}. 
Recent studies also leverage the label or pseudo label as guidance to increase precise of pair construction~\cite{zhao2020contrastive,wang2021exploring}. 
However, they still only address semantic segmentation of a single image.
Our method is inspired by this stream with merit of more reliable contrast, but we establish a different pair construction tailored for surgical video analysis, \ie, extending the pair selection range to different videos, which enables global constraints on a wider embedding space.
\textcolor{black}{Moreover, model optimization is another key factor in contrastive learning with several typical architectures standing out, such as SimCLR using all samples within the current mini-batch for end-to-end update~\cite{chen2020simple}; MoCo using momentum-updated encoder to encode samples on-the-fly for learning representations~\cite{he2020momentum}; BYOL also using momentum update but avoiding relying on negative pairs, which is superior when negative pair is hard to be defined~\cite{grill2020bootstrap}.
Each stream inspires many following works according to the tasks they aim to tackle~\cite{li2020prototypical,xiao2020should,xie2021propagate}.
\textcolor{black}{In our work, we propose to leverage ground truths to guide the precise construction of negative pairs. Therefore, we design our contrastive method following MoCo stream, to borrow the information from negative pairs via symmetric architecture.}
We further present difference in loss objective function and training scheme, by considering the properties of pixel-level scene segmentation.
}

\section{Methodology}

Fig.~\ref{fig:overview} presents an overview of our proposed \emph{STswinCL} for whole scene segmentation of surgical video.
In this section, we first introduce the overall architecture of our developed Transformer-based model.
Then we describe the joint space-time shift window scheme in Transformer, to efficiently relate the richer cues to augment each pixel of the current video frame.
We next illustrate the pixel-to-pixel contrast via a new label selected multi-source pair construction, to capture even wider relations across different videos.
Overall training objective of these two complementary factors, and collaboratively training procedure are described finally.

\vspace{-1mm}
\subsection{Overall Architecture} 
\label{sec:overall}
To generate the feature of frame $x_t$, we extract a video clip as the model input, which contains the current frame $x_t$ and a set of its previous frames, as $\bm{x} \! = \! \{ x_{t-m}, \ldots, x_{t-1}, x_t \}$, with $N$ frames.
The frames in the clip first pass forward a CNN backbone to generate corresponding feature maps $\bm{f}\! = \! \{ f_{t-m}, \ldots, f_{t-1}, f_t \}$.
$f_t \!\in\! \mathbb{R}^{\frac{H}{S} \!\times\! \frac{W}{S} \!\times\! C}$, where $C$ is the number of channels, $S$ is the sampling stride, and $H\!\times\! W$ is the frame size. 
\textcolor{black}{For pixel $i$ in $x_t$, its corresponding embedding} in the feature map $f_t$ is denoted as $f_t^i$, and we consider each pixel embedding as a token.
Transformer blocks with proposed self-attention scheme (STswin in Sec.~\ref{sec:trans}) are then employed on each token, to enhance the representation capability by borrowing the information from other tokens. The number of tokens is maintained as $\frac{H}{S} \!\times \!\frac{W}{S}$ with the channel as $C$.
To encode multi-scale hierarchical representations, we then explore a feature merging layer as the model goes deeper. It concatenates the features of $2 \!\times\! 2$ neighboring tokens to enlarge the reception field of each token.
A linear projection layer is followed to reduce feature channel, obtaining a feature map with resolution $\frac{H}{2S}\! \times \!\frac{W}{2S} \!\times\! 2C$. STswin Transformer blocks are employed afterwards for feature transformation.
The merging procedure can be repeated multiple times (empirically performing once in this work).
ASPP~\cite{chen2017deeplab} and feature concatenation~\cite{chen2018encoder} strategies are then used for integrating feature information from different scales and extraction levels,
finally outputting feature maps with resolution $\frac{H}{S} \!\times\! \frac{W}{S}  \!\times\! D_\text{Tr}$ into different heads for model optimization by different objectives (Sec.~\ref{sec:contrast}).
Note that unlike Transformer in~\cite{liu2021swin} purely relying on self attention, our model shows a hybrid CNN-Transformer form, leveraging CNN with strong inductive biases for feature extraction, which can avoid model overfitting on small dataset of surgical video.

\vspace{-1mm}
\subsection{Intra-video Relating with STswin Transformer}
\label{sec:trans}
\textbf{Self-attention within local window.}
The standard Transformer conducts self-attention between a token and all other tokens in a image, leading to quadratic complexity with respect to the token number. It is intractable for dense scene segmentation with \textcolor{black}{a large amount of} tokens.
Sharing the similar spirit with~\cite{liu2021swin}, we propose only to compute self-attention within the local window.
We evenly partition the feature maps extracted from the CNN backbone to several non-overlapping windows, with feature maps at different timesteps partitioned in the same way (cf. Fig.~\ref{fig:overview}(a), partitioning via yellow dashline on a cropped map for concise illustration).
We then only attend the internal tokens to each other \textcolor{black}{within the current local window in (a)}, \ie, for a yellow token, we only use green tokens in (a).
Different from Transformer in \cite{liu2021swin}, that performs window partition on input image, our hybrid CNN-Transformer conducts window partition on higher-level feature maps extracted from CNN backbone.
Moreover, our window is expanded to the temporal dimension, yet also with a local range (only two timesteps in our work).
The window size keeps stable, 
resulting in a fixed number of tokens in each window.
\textcolor{black}{Therefore, computational complexity can substantially decrease compared with the standard Transformer when modeling long temporal cues of large size images, making window based self-attention feasible for dense prediction of videos.
It also avoids involving wrong relations among similar classes at completely different (distant) locations.}

\textbf{Window interaction with space-time shift.}
Local window with relatively less information may limit the feature discrimination power.
To induce a cross-window connection while retaining efficient computation, we explore a joint space-time shift by alternating the different configurations for Transformer blocks and arranging them in a consecutive way.
As illustrated in Fig.~\ref{fig:overview}(a), the regular window partitioning strategy starts from the top-left \textcolor{black}{pixel embedding}, and the $4 \!\times \! 4$ feature map is evenly divided into $2 \!\times\! 2$ windows of size $2 \!\times\! 2$  ($M=2$ denoting the window size).
It is followed by another partitioning configuration acting as \textbf{\emph{space shift}} in (b): rolling the windows by ($\lfloor \frac{M}{2}\rfloor, \lfloor \frac{M}{2}\rfloor$) \textcolor{black}{pixel embeddings} from preceding partitioned ones.
These two settings are arranged as two consecutive layers forming a Transformer block, which transforms richer cues (including green tokens in (b)) for a token (yellow token).
The consecutive computation in a Transformer block is defined as:
\vspace{-3mm}
\begin{equation}
\label{eq:overall}
\begin{split}
& \hat{f}^l_t = \mathcal{T}^\text{R}( \text{LN}(f^{l-1}_t), \text{LN}(f^{l-1}_{t-1})) + f^{l-1}_t, \\
& f^l_t = \mathcal{M}( \text{LN}(\hat{f}^l_t)) + \hat{f}^l_t, \\
& \hat{f}^{l+1}_t = \mathcal{T}^\text{S}( \text{LN}(f^{l}_t), \text{LN}(f^{l}_{t-1})) + f^{l}_t, \\
& f^{l+1}_t = \mathcal{M}( \text{LN}(\hat{f}^{l+1}_t)) + \hat{f}^{l+1}_t,\\
\end{split}
\vspace{-5mm}
\end{equation}
where taking the output feature $f^{l-1}_t$ of frame $x_t$ from layer $l\!-\!1$ as input, window based Transformer layers in regular and shifted window partitioning configurations ($\mathcal{T}^\text{R}$ and $\mathcal{T}^\text{S}$) are employed successively, \textcolor{black}{which are similar to~\cite{liu2021swin} but space shift are performed on 2-timestep frames in the same way.}
MLP module ($\mathcal{M}$) is followed afterwards each Transformer layer. LN represents LayerNorm.

Space shift allows each token receiving broader view from neighbor pixels, while the information in temporal dimension is still limited to the local window, \ie, 2 timesteps.
One direct way to extend time cues is sampling input frames at a interval, yet easily losing the crucial adjacent information.
Instead, we explore \textbf{\emph{time shift}} to extend the temporal view to longer scope. 
Taking a 4-timestep video clip as an example, it is achieved by shifting time index into one Transformer block (cf. Fig.~\ref{fig:overview}(c)).
Coupling $\{f_t, f_{t-1}\}$ and coupling $\{f_{t-2}, f_{t-3}\}$ into respective blocks, only the information in $t\!-\!1$ can flow to the current frame pixel $t$. \textcolor{black}{Building up on the update features,} we then shift the time index by coupling $\{f_{t-1}, f_{t-2}\}$ into one block to let frames $t\!-\!2$ affect the frame $t\!-\!1$, with implicit interaction from $t\!-\!3$. Shifting back to the original time configuration, $f_t$ can capture the temporal cues of 4 timesteps ago, updated to:
\begin{equation}
\label{eq:overall}
\begin{split}
& \textcolor{black}{\hat{f}^{l+5}_t = \mathcal{T}^\text{S}( \text{LN}(f^{l+4}_t), \text{LN}(f^{l+4}_{t-1}) ~|~ f^{l+4}_{t-3:t}) + f^{l+4}_t,} \\
& f^{l+5}_t = \mathcal{M}( \text{LN}(\hat{f}^{l+5}_t)) + \hat{f}^{l+5}_t.\\
\end{split}
\end{equation}
Note that the same scheme can be readily performed on a longer clip for modelling longer temporal cues.
\textcolor{black}{
And it shall follow two principles: i) we enforce only 2 timesteps into one block, to make the most relevant information from the adjacent frame explicitly flow to each frame, while keeping the implicit communication with other frames.
ii) the time shift with displacing the time index of one is performed until each frame in the clip can capture the complete temporal information from all the other frames in the clip.
As it is essential for our video scene segmentation task, to facilitate model to predict accurate segmentation mask for every frame.
In this regard, the longer input clip requires more time shifts (e.g., 8-timestep clip needs 6 time shifts). 
However, compared with the standard Transformer that requires the attention operations between each frame with all the other frames, the saved computational cost attributed to our time shift also enlarges as the clip length increases.}
Cyclic shift with a masking mechanism~\cite{liu2021swin} is also employed for the efficient batch computation.

\vspace{-1mm}
\subsection{Inter-video Relating with Supervised Pixel Contrast}
\label{sec:contrast}
\textbf{Pixel-level cross-entropy loss.}
Our STswin Transformer enhances the feature representation capacity of frame $x_t$, obtaining feature map $z_t$ by seeing rich views.
In the context of scene segmentation, each pixel $i$ of $x_t$ needs to be classified into a semantic class $c \in \mathcal{C}$.
We first exploit the standard pixel-level cross entropy (CE) loss for optimization.
The feature map $z_t$ is upsampled to produce dense embeddings and then mapped to a categorical score map by the segmentation head: $\mathcal{Y}_t = \mathcal{F}_{\text{SEG}}(\text{Up}(z_t)) \in \mathbb{R}^{H \times W\times |\mathcal{C}|}$. 
Denoted score embedding for pixel $i$ as $\mathcal{Y}_t^i$, and ground-truth label as $y_t^i \in \mathcal{C}$, the pixel-level CE loss is calculated as:
\begin{equation}
\label{eq:CE}
\mathcal{L}_i^{\text{CE}} = -\textbf{1}_{y_t^i}^{\top} \text{log}(\text{Softmax}(\mathcal{Y}_t^i)),
\end{equation}
where $\textbf{1}_{y_t^i}$ represents the one-hot encoding of $y_t^i$. 

\textbf{Inter-video pixel-to-pixel contrast.}
Existing methods ignore capturing inter-video relation, because the purely used CE loss can only independently induce pixel-to-groundtruth regularization, rather than pixel-to-pixel comparison across videos.
Global embedding space shows ambiguous boundary, which hardly tackles the high pixel similarity from different classes yet low pixel similarity within the same class.
In addition, inherent label bias in the imbalanced surgical data drastically alters the decision boundary by majority classes.
We therefore explore the inter-video relation by developing the pixel-level contrastive learning as an additional constraint.
Global embedding space can be further regularized to a better structure for accurate segmentation.

Given a frame $x_t \! \in \! \mathbb{R}^{H\!\times \!W \!\times\! 3}$ as the query sample and a set of frames as key samples, image-level contrastive learning trains the model by distinguishing the positives (augmentation version of $x_t$) from negatives (from key samples excluding $x_t$)~\cite{he2020momentum}.
Our pixel-level contrastive learning extends data sample from image to pixel.
Here, we analyze pixels in current frame $x_t$ as the query samples and omit $t$ for brevity. 
Concretely, our contrast duplicates the model into two symmetric encoder branches, including a regular branch for encoding query sample that is updated online via back propagation, and a momentum branch for key samples which is updated slowly by exponential moving average (EMA)\cite{hunter1986exponentially}.
Each branch consists of the developed Transformer and a projection head $\mathcal{F}_{\text{PROJ}}$ to map features to a lower dimension.
We feed current frame $x$ into the regular encoder branch to predict the query feature map. Then we select the key samples from three types of sources (cf. Fig.~\ref{fig:overview}) to model global relations:
1) the same frame with a different perturbation; 2) adjacent frame from the same video $\bm{v}$ which mainly contributes to positive pairs given the similar visual context; 3) frame from other videos $\bm{\tilde{v}}$ which mainly increases negative pairs.
Note that more positives and negatives also benefit contrastive learning~\cite{chen2020simple}. 
All these key frame samples are fed into a momentum encoder branch to predict the key feature maps.
Denoting query and key feature maps as $e_q,e_k \! \in \! \mathbb{R}^{\frac{H}{S} \! \times \! \frac{W}{S} \!\times\! D_{\text{cl}}}$, we extract pixel-level feature embeddings $e_q^i$ and $e_k^j$ for the $i$ and $j$ pixels, respectively.

Most existing works relied on the spatial distance for pixel pair construction~\cite{xie2021propagate}, 
which is unreliable and even infeasible for pixels from different videos. We leverage the ground-truth label for pair selection under the assumption that the embeddings of pixels belonging to the same class should be closer than those from different classes. 
We perform the same augmentation on labels and do the resolution reduction to $y_q,y_k \! \in \! \mathbb{R}^{\frac{H}{S} \! \times \! \frac{W}{S} \!\times\! |\mathcal{C}|}$, then couple $\{e_q^i,y_q^i\}$ for query pixel $i$ and $\{e_k^j,y_k^j\}$ for key pixel $j$.
For the query pixel $i$, we derive a label selection mask for each key feature map $e_k$, with a binary variable: $\mathcal{M} \subseteq \{0,1\}^{\frac{H}{S} \! \times \! \frac{W}{S}}$. The value corresponds to pixel $j$ is determined by the label:
\begin{equation}
\vspace{-1mm}
\label{eq:mask}
\mathcal{M}_{j} = \mathds{1}[{y_q^i = y_k^j}],
\vspace{-1mm}
\end{equation}
where $\mathds{1}(\cdot)$ is the indicator function. $e_k^j$ is categorized as positive sample when two pixels belong to the same class ($\mathcal{M}_{j}\! = \!1$), denoting as $e_k^{j^+}$. Otherwise, it falls into negative sample as $e_k^{j^-}$.
Our pixel-level contrastive objective is then defined to maximize the similarities of positive pairs while minimizing the similarities of negative pairs:
\begin{equation}
\label{eq:CL}
\small
\vspace{-2mm}
\mathcal{L}_i^{\text{CL}} =  - \log \frac{\exp (S_i^p)}{\exp (S_i^p)+ \exp (S_i^n)},
\end{equation}
\begin{equation}
\label{eq:similarity}
\footnotesize
S^p_i = \frac{1}{| \mathcal{P}_i|} \!\sum_{j^+ \in \mathcal{P}_i} \! \langle e_q^i \cdot e_k^{j^+}\rangle, ~~ S^n_i = \sum \! \frac{1}{| \mathcal{N}^n_{i}|} \! \sum_{j^- \in \mathcal{N}^n_{i}} \! \langle e_q^i \cdot e_k^{j^-} \rangle,
\vspace{-2mm}
\end{equation}
where $\mathcal{P}_i$ and $\mathcal{N}_i$ are pixel embedding collections of positive and negative samples for pixel $i$, \textcolor{black}{and we include all the samples that successfully form as positive and negative pairs with $i$ in current mini-batch. Meanwhile, as our pixel-level contrast can construct the abundant pixel pairs in each mini-batch given each frame containing lots of pixels, we omit the queue strategy in MoCo~\cite{he2020momentum} in model optimization.}
$\langle \cdot \rangle$ denotes \textcolor{black}{cosine similarity function.} 
\textcolor{black}{
In our pixel-level contrastive learning, the amount of positive and negative pairs is unfixed for different pixels during different training iterations, 
therefore, we perform the normalization on feature similarity.
Note that we propose to normalize the negative similarity in frame-level,} \ie, within the subset of key pixels from one frame: $\mathcal{N}^n_{i} \! \subset \! \mathcal{N}_i$.
Compared with the positive one that is averaged among all pixels, this strategy can enlarge the value of negative proportion to benefit model learning.
In addition, in Eq.~\ref{eq:similarity}, $\mathcal{P}_i$ and $\mathcal{N}_i$ are derived from the cross-video key frames of three-type sources which is not restricted to the same video as pixel $i$.
To this end, 
our pixel-to-pixel contrast in Eq.~\ref{eq:CL} can capture the relation between pixels from whole data and explore the global structure of embeddings.

\vspace{-3mm}
\subsection{Learning with Dual Relations}
\subsubsection{Objective Function}
The pixel-level CE loss in Eq.~\ref{eq:CE} aims to learn the discriminative pixel features for meaningful classification.
The unary form lets the model learning focus on increasing the feature capacity: capturing rich spatial and temporal relation from the current video via our STswin Transformer.
Complementarily, our pixel-to-pixel contrast loss in Eq.~\ref{eq:CL} helps to explicitly explore a more global semantic correlation between pixel samples \textcolor{black}{across} videos for a well-shaped embedding structure. 
Our overall training objective includes both: $\mathcal{L} = \mathcal{L}^{\text{CE}} + \mathcal{L}^{\text{CL}}$, and we equip CE loss with online hard example mining strategy~\cite{shrivastava2016training}.

\subsubsection{Network Configuration and Training Procedure}
We utilize DeepLabV3+~\cite{chen2018encoder} as our backbone model with a ResNet-18 encoder pre-trained on ImageNet~\cite{he2016deep}.
The segmentation head $\mathcal{F}_{\text{SEG}}$ is composed of two convolutional layers with kernel size $3 \!\times\! 3$ and $1 \!\times\! 1$ for dense classification.
The projection head $\mathcal{F}_{\text{PROJ}}$ consists of two $1 \!\times\! 1$ convolutional layers to reduce feature channels.
We employ a three-stage training strategy to train our model. 
\textcolor{black}{The inputs for all three stages are a number of video clips selected from different videos, with each clip containing continuous video frames.}
We first train our Transformer model with segmentation head, \textcolor{black}{outputting the predicted segmentation masks. The model is trained towards ground truth masks by cross entropy loss,} to provide a good initialization of embedding space.
Next, we perform our inter-video contrastive learning by replacing the segmentation head with projection head, with other parameters initialized by the model trained in stage one. \textcolor{black}{The outputs are the extracted feature embeddings, which are optimized to be pulled close or pushed apart according to the class labels, using the contrastive loss.}
Finally, we perform the standard training procedure by equipping the segmentation head back, to fine-tune the model purely towards cross entropy loss.
Note that we learn the model towards the contrast objective in a pre-training manner, instead of jointly training the model with the cross entropy loss. We also empirically find that this training strategy is better than the joint training version.
The underlying reason is that 
it can purify the model training to tackle the class imbalance in surgical video.
The label bias in the imbalanced data can be alleviated, by avoiding too heavy constraint when using cross entropy loss to directly regularize the label information onto embeddings.
Instead, pre-training manner can learn a better initialization with more label-agnostic from the imbalanced data.
During inference, we sequentially create the input frame clip from each video in the form of a sliding window, with each time shifting one frame forward.
It forwards to Transformer model with segmentation head to produce final prediction mask.

\section{Experiments}


\subsection{Datasets and Implementation}

\textbf{EndoVis18.} We use a public challenge dataset from 2018 MICCAI Robotic Scene Segmentation Challenge, referred to as EndoVis18~\cite{allan20202018}. This dataset consists of 19 sequences, officially divided into 15 for training and 4 sequences for testing. Each training sequence contains 149 frames while each testing sequence has 249/250 frames. 
All sequences are recorded on da Vinci X or Xi system during porcine training procedure. 
Each frame is of a high resolution of 1280$ \times $1024. 
The dataset parses the entire surgical scene into 12 classes, including different anatomy and robotic instruments.

\textbf{CaDIS.} 
We utilize a public benchmark dataset CaDIS, focusing on cataract surgery~\cite{grammatikopoulou2021cadis}. 
It consists of 25 videos, officially split into 19, 3 and 3 videos as training, validation and testing datasets. 
Videos are recorded using a 180I camera mounted on an OPMI Lumera T microscope and each frame has a high resolution of 960$ \times $540.
The dataset provides pixel-level annotations for anatomical structures, instruments and miscellaneous objects, and three sub-tasks are defined with increasing granularity. 
Task I involves 8 classes, including 4 anatomies, 3 miscellaneous classes and 1 class for all the instruments.
Take II incorporates instrument classification and increases class number to 17. The instruments are grouped in categories according to appearance similarities, resulting in 9 distinct classes.
Task III allows more granular instrument classification by representing each instrument type to tips and handles as separate classes, obtaining 25 classes overall.
The classes that do not appear in all data splits and are present in less than 5 videos are ignored during training.

For fair comparison, we exactly follow the original evaluation protocol of respective datasets \cite{allan20202018,grammatikopoulou2021cadis}.
For EndoVis18 challenge, we use mean intersection-over-union (mIoU) to evaluate methods. We also add one commonly-used metric, \ie, Dice coefficient (Dice) to comprehensively validate the performance.
The score for each frame is first calculated, in which only the classes present in the given frame are counted and the background class is excluded.
We then average scores over all frames to get the final results.
Overall and individual results on four test sequences are reported. 
For CaDIS benchmark, mIoU, Pixel Accuracy per Class (PAC), and Pixel Accuracy (PA) are employed for assessment.
Ignored classes are not taken into account when the metrics are calculated.
More detailed definition can be referred to \cite{allan20202018,grammatikopoulou2021cadis}. 

\textbf{Implementation Details.}
Our framework is implemented based on PyTorch using 2 NVIDIA RTX 3090 GPUs for training. 
All video frames are resized to the resolution of $640 \!\times \! 512$ in stage one and three.
The data augmentations with cropping, random scaling and rotation are performed to enlarge the training dataset.
We deploy SGD optimizer with a poly learning rate scheduler and use a base learning rate of $1e\!-\!3$ for these two stages.  
Batch size is set to 8 and the clip length is empirically set to 4.
In contrastive learning, we randomly crop a patch from a frame (with varying resolutions) and then resize it to $448 \!\times \!256$ to create more aggressive augmentation. 
We use LARS optimizer and a cosine learning rate scheduler with initial learning rate as 1. 
Batch size is set to 4 and the clip length is also set to 4 for consistency with other two stages. 
We include one adjacent frame and three frames from other videos into contrastive learning. 
\textcolor{black}{We randomly select three videos from the training dataset, and from each video, we randomly sample one frame to form these three frames.
To determine the training epoch of contrastive learning, we select the model of that enables the best segmentation performance in stage three. 
}


\vspace{-2mm}
\subsection{Comparison with State-of-the-arts}

\subsubsection{EndoVis18}
We compare our method with state-of-the-arts, first including top-five approaches reported in challenge. Both IRCAD and Digital are based on DeepLabV3+~\cite{chen2018encoder}, and improve results by aggressive data augmentation and auxiliary prediction.
NCT develops method based on TernausNet~\cite{shvets2018automatic}, a skip-connection model trained with transfer learning.
UNC~\cite{ren2020task} decomposes the single task into three subsequent sub-tasks at different perceptual levels, ranking the second place. 
OTH utilizes massive data augmentation and leverages extra data from EndoVis17 Challenge~\cite{allan20192017} to train DeepLabV3+, winning the \textcolor{black}{championship}.
All these methods rely on visual cues from a single frame.
We then include the latest study on this dataset (Noisy-LSTM)~\cite{wang2021noisy} for comparison, which uses ConvLSTM to learn temporal coherence. 

Results of different methods are presented in Table~\ref{tab:result}.
UNC is from the original paper~\cite{ren2020task} and the rest in challenge are quoted from~\cite{allan20202018}.
Noisy-LSTM only reports results on their own split validation data, we therefore re-implement the method using their released code and list results on test data for fair comparison.
\textcolor{black}{We see that compared with methods solely using a single frame on the current dataset without leveraging extra data, i.e., UNC,} our STswinCL gains great improvement by 2.9\% mIoU, raising $60.7\%$ to $63.6\%$, indicating that richer intra-video relation can increase feature discrimination.
STswinCL also largely outperforms the latest time modelling method Noisy-LSTM by 3.2\% mIoU and 2.9\% Dice, showing the importance of inter-video relation modelling to better structure feature space.
Although OTH includes additional label information, our STswinCL still achieves superior overall results and results on most sequences, peaking a new state-of-the-art of $63.6\%$ mIoU and $72.0\%$ Dice.

\subsubsection{CaDIS}
We further compare our method with state-of-the-arts in CaDIS, \ie, three well-known and cutting-edge networks for whole scene segmentation task, including DeepLabV3+ with lightweight version~\cite{chen2018encoder}; UPerNet~\cite{xiao2018unified} (a multi-task framework to discover the rich knowledge by parsing multiple visual
concepts at once); and more advanced one HRNetV2~\cite{wang2020deep} (preserving the high-resolution feature representations by integrating multi-scale context). 
We quote their results reported in~\cite{grammatikopoulou2021cadis} and list all the results in Table~\ref{tab:cadis}.
It is observed that HRNetV2 largely outperforms the DeepLabV3+ network, showing strong performance on all three tasks.
Our proposed STswinCL achieves superior performance to HRNetV2 across all the evaluation metrics of all the three tasks, peaking the new state-of-the-arts.
Our mIoU consistently outperforms HRNetV2 by 0.8\%, 1.2\%, and 3.1\% at the three tasks, where the larger gains are obtained as the challenge of the task increases with more fine-grained granularity.
The improvement of PAC is also obvious especially on the most challenging task (Task III), boosted from $77.0\%$ to $78.6\%$.
\begin{table}[!t]
\caption{\textcolor{black}{Segmentation results of different methods \\ on EndoVis18 dataset (12 Classes). }}
\vspace{-2mm}
\label{tab:result}
\centering
\resizebox{0.48\textwidth}{!}{
\begin{threeparttable}
\begin{tabular}{c|cccccc}
\hline
  \multirow{2}{*}{Method} &
  \multirow{2}{*}{mIoU(\%)} &
  \multicolumn{4}{c}{Sequence (mIoU(\%))} &
    \multirow{2}{*}{Dice(\%)} \\ \cline{3-6} 
                    &   & Seq1    & Seq2    & Seq3   & Seq4     &     \\ \hline
  IRCAD~\cite{chen2018encoder}    & 57.3 & 68.8 & 52.9 & 79.0 &  28.5  &-  \\
  Digital~\cite{chen2018encoder}  & 57.9 & 63.6 & 54.9 & 80.6 &  32.4  &- \\
NCT~\cite{shvets2018automatic}      & 58.5  & 65.8 & 55.5 & 76.5 &  36.2  &-\\
UNC~\cite{ren2020task}         & 60.7 & 63.3 & 57.8 & 81.4&  37.3  &- \\ 
$\text{OTH}^{\bm{\ast}}$~\cite{chen2018encoder}       & 62.1 & \textbf{69.1}& 57.5 & 82.9 & 39.0  &- \\ \hdashline
Noisy-LSTM~\cite{wang2021noisy}     & 60.4 & 67.0 & 56.3 & 81.8 &36.4   & 69.1\\ 
Swin~\cite{liu2021swin} & 60.8 & 66.8 & 57.1 & 81.9 & 37.3   & 69.9 \\
VideoSwin~\cite{liu2021video}  & 57.9  & 62.2  & 58.9 & 81.2& 29.3   & 66.6\\
\hline 
Baseline   & 61.2 &  65.3 & 56.4 &  83.4 & 39.9   &  70.2 \\ 
BaselineCL   & 61.6 &  63.8 & 58.8 &  81.7 & \textbf{41.8}   &  70.6 \\ 
Sswin     &  61.7 & 63.8 & 59.7 & 83.2 & 40.0   & 70.4 \\ 
STswin     & 62.3 &  65.6 & 60.9 &  \textbf{84.3} & 38.5   &  70.8  \\ \hline
\textbf{STswinCL (ours)}   & \textbf{63.6} &  67.0 & \textbf{63.4} &  83.7 & 40.3   &  \textbf{72.0} \\ \hline

\end{tabular}
\end{threeparttable}
}
\vspace{1mm}
\scriptsize Note: $\bm{\ast}$ means the method use extra training data. ~~~~~~~~~~~~~~~~~~~~~~~~~~~~~~~~~
\vspace{-7mm}
\end{table}

\begin{table*}[]
\centering
\caption{\textcolor{black}{Results of different methods on CaDIS dataset for scene segmentation.}}
\vspace{-2mm}
\begin{tabular}{c|ccc|ccc|ccc}
\hline
\multirow{2}{*}{Methods} & \multicolumn{3}{c|}{Task I (8 Classes)} & \multicolumn{3}{c|}{Task II (17 Classes)}  & \multicolumn{3}{c}{Task III (25 Classes)} \\ \cline{2-10}
& mIoU(\%)                  & PAC(\%)                    & PA(\%)      & mIoU(\%)                  & PAC(\%) & PA(\%)                   & mIoU(\%)                   & PAC(\%)                    & PA(\%)          \\ \hline

DeepLabV3+ \cite{chen2018encoder}   & 82.6   & 88.7  & 93.9   & 72.3    & 80.8  & 93.5  & 63.2    & 75.6   & 93.9   \\
UperNet \cite{xiao2018unified}     & 84.0   & 89.5  & 94.2   & 73.8    & 82.0  & 94.1  & 66.8    & 77.8   & 94.2   \\
HRNetV2 \cite{wang2020deep}        & 84.9    & 90.0  & 94.2  & 76.1    & 83.6  & 94.6  & 66.6    & 77.0  & 94.3    \\ \hdashline
Swin~\cite{liu2021swin}  & 84.4    & 90.3  & 94.2  & 73.8    & 84.2  & 94.0  & 62.1    & 74.9  & 94.1    \\
VideoSwin~\cite{liu2021video}  & 82.8    & 89.8  & 93.9  & 71.2    & 82.0  & 93.8  & 64.5    & 76.2  & 93.9    \\
\hline
Baseline                 & 84.4   & 90.2 & 94.3    & 73.9   & 81.2 & 94.0    & 64.0   & 72.3 & 93.8 \\
BaselineCL                 & 84.8   & 90.5 & 94.5    & 74.4   & 81.9 & 94.3    & 64.0   & 73.4 & 94.2 \\
Sswin                     & 85.0   & 90.3 & 94.4   & 74.9    & 82.7 & 94.0  & 66.4   & 74.4   & 94.2   \\ 
STswin             & 85.2   & 90.5  & 94.5   & 75.7    & 84.3  & 94.1  & 68.0    & 77.3 & 94.3 \\ \hline
\textbf{STswinCL (ours)}       & \textbf{85.7}   & \textbf{91.4} & \textbf{94.6}    & \textbf{77.3}    & \textbf{84.5} & \textbf{94.6}   & \textbf{69.7}   &  \textbf{78.6} & \textbf{94.5}  \\ \hline
\end{tabular}
\label{tab:cadis}
\end{table*}

\begin{figure*}[t]
	\centering
	\includegraphics[width=0.96\textwidth]{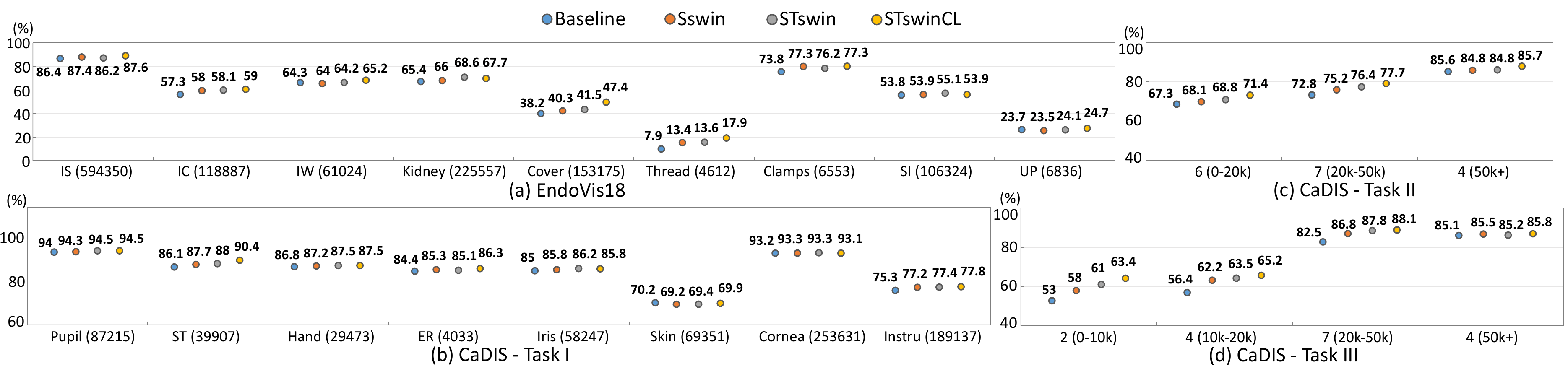}
	\vspace{-3mm}
	\caption{\textbf{Class-level IoU} of four settings.
		Results of all classes in EndoVis18 \& Task I are shown: class name (number of pixels per frame). We group classes in Task II \& III for clear comparison: the number of classes (pixel ranges).
		Average results of test videos are shown by solid circles.}
	\label{fig:bar}
	\vspace{-2mm}
\end{figure*}

\vspace{-2mm}
\subsection{Effectiveness of Key Components}

We conduct ablation experiments to validate the effectiveness of different key components in the proposed method and obtain four configurations:
(i) \textit{Baseline}: we train the pure DeepLabV3+ backbone network as the baseline;
(ii) \textit{Sswin}: we train our Transformer model with self-attention within local window, and perform \textbf{s}hifted \textbf{win}dow only in \textbf{S}patial dimension with single frame as input; 
(iii) \textit{STswin}: we form the video sequence as input to train our Transformer, and both self-attention region and shifted window are extended to the joint \textbf{S}pace-\textbf{T}ime dimensions, while training objective is the pure $\mathcal{L}_i^{\text{CE}}$ in Eq.~\ref{eq:CE};
\textcolor{black}{
(iv) \textit{BaselineCL}: we add our inter-video supervised \textbf{C}ontrastive \textbf{L}earning $\mathcal{L}_i^{\text{CL}}$ in Eq.~\ref{eq:CL}, in combination with $\mathcal{L}_i^{\text{CE}}$, to train the Baseline model.
(v) \textit{STswinCL}: we include our inter-video \textbf{C}ontrastive \textbf{L}earning to STswin, to complete our proposed model.}

The results on EndoVis18 and CaDIS are presented in Table~\ref{tab:result} and Table~\ref{tab:cadis}, respectively.
We observe that DeepLabV3+ backbone obtains reasonable results with mIoU over 60\% on all tasks of both datasets.
Purely introducing self-attention mechanism on a spatial dimension, Sswin Transformer achieves better results.
Incorporating temporal coherence cues, STswin further improves the segmentation performance in all evaluation metrics of both datasets.
Especially, mIoU on EndoVis18 is increased from 61.7\% to 62.3\%, and PACs of Task II \& III on CaDIS are increased from 82.7\% to 84.3\%, and from 74.4\% to 77.3\%.
Adding our cross-video contrast loss, our full model STswinCL continues \textcolor{black}{boosting the results with 0.5\%-1.7\% mIoU gain,} peaking at 63.6\% mIoU on EndoVis18 and 69.7\% mIoU on CaDIS Task III. 
\textcolor{black}{Introducing our contrast loss to Baseline model, BaselineCL can also consistently raise the results on both datasets, especially with 0.7\% and 0.9\% PAC gain in CaDIS Task II and III.}
Note that when tackling challenging tasks with more categories (\eg, comparing Task I and III on CaDIS), our key components can instead yield their better efficacy with larger improvement.

To validate the effectiveness of proposed hybrid architecture of CNN and Transformer, and also time shift scheme, we further compare our method with i) Swin: the original Swin Transformer without any CNN~\cite{liu2021swin}, and ii) VideoSwin: the video version, which is originally designed for video action recognition and only performs the shift once in most feature encoding stages~\cite{liu2021video}.
We reimplement these methods using their released codes, use the pretrained models and utilize DeepLabV3+ as the base segmentation framework for fair comparison.
We try the best to adapt VideoSwin to achieve the better results for segmentation by trying different down-sampling rates in feature extraction, different learning rates in model learning, as well as different ways to represent the feature for current frame.
From Table I and II, we can see that the original Swin method is inferior to our Sswin on most evaluation metrics. The performance of original Swin is even worse than Baseline model (ResNet-18) on EndoVis18.
Additionally, apart from PAC of Task I and III on CaDIS, there exists large performance gap between VideoSwin and our STswin.
These results demonstrate that without using CNN with strong inductive biases for feature extraction, pure Transformer is prone to the model overfitting when using small surgical dataset for training.
Our CNN-Transformer hybrid architecture can alleviate this issue.
In addition, shifting once hardly captures the complete temporal cues within the clip. It may be sufficient for action recognition task, while is suboptimal for our scene segmentation task that requires a dense pixel-wise prediction for each frame.
The performance of VideoSwin is even lower than Swin. The reason may be that by extending Transformer to the temporal dimension, VideoSwin introduces more model parameters, which further increases the severity of overfitting problem.
Evaluating and comparing on a larger dataset can better verify the effectiveness of our method.




\subsubsection{Class-level Comparison} 

Surgical scene often exhibits the heavy class imbalance, posing a great challenge for segmentation models.
To validate the capacity of different components in the context of imbalanced learning, we illustrate charts to show the detailed results in class-level in Fig.~\ref{fig:bar}.
For EndoVis18 \& Task I, numerical results on all the classes are shown. 
While for Task II \& III with massive class numbers, we first group classes according to their pixel amount and then average results over classes within each group, for clearly comparing the efficacy of our methods on different rareness levels. 
We see that both joint space-time relation (STswin) and pixel-to-pixel relation (STswinCL), can achieve consistent improvements in most classes across both datasets.
Compared with Sswin, STswin increases performance of different classes by a similar degree.
While STswinCL, as we expected, demonstrates its great effectiveness especially in the minor classes, given well-structured embedding space.
For example, Thread and ER in Task I achieves noticeable gain of 4.3\% and 1.2\%, respectively. 
The smallest portions in Task II\&III also show the most significant increase by 2.6\% and 2.4\%.

\begin{figure*}[!ht]
	\centering
	\includegraphics[width=0.78\textwidth]{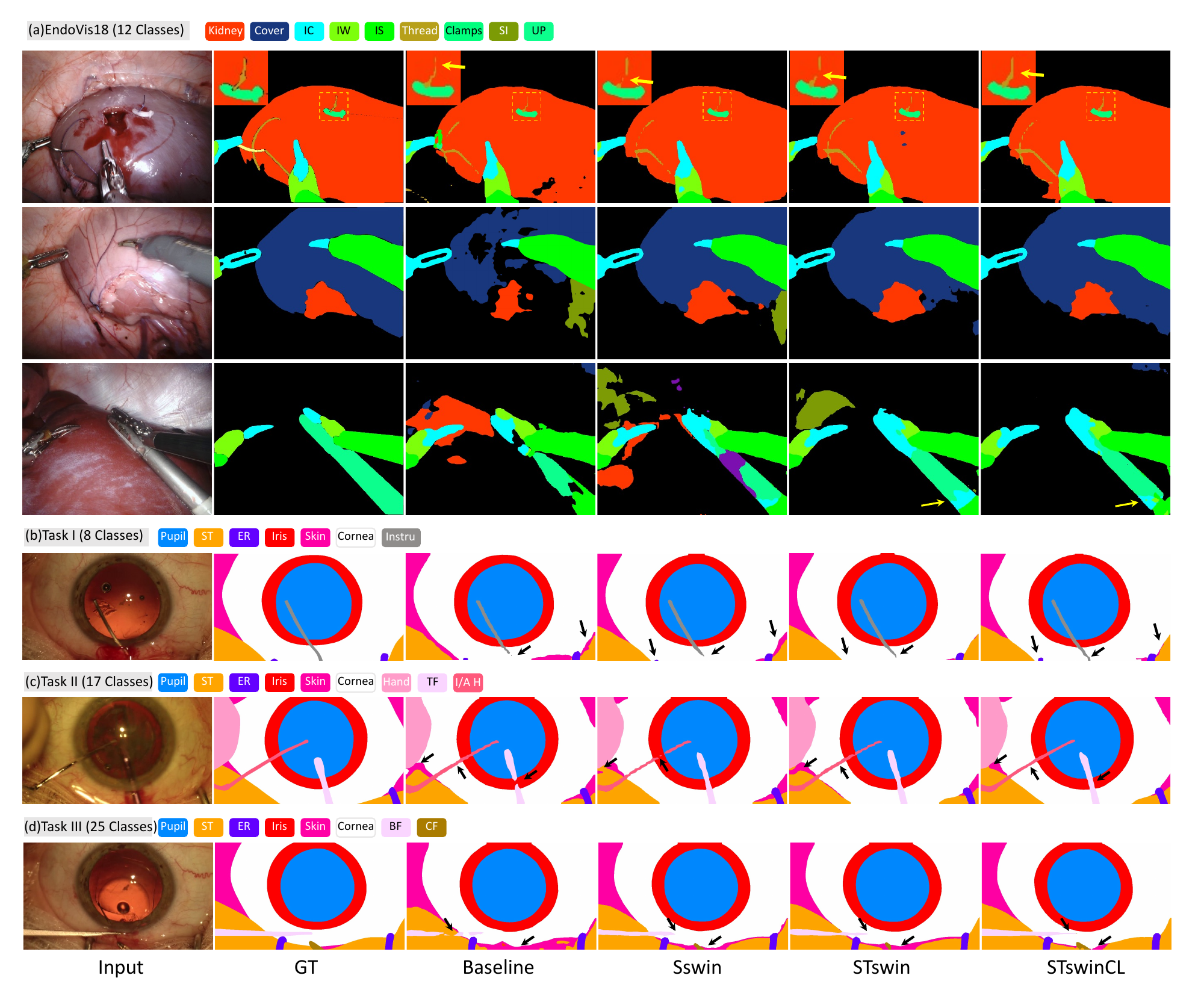}
	\vspace{-3mm}
    	\caption{\textbf{Qualitative comparisons} of different methods on the typical frames from (a) EndoVis18 and (b-d) CaDIS datasets. Different color represents different semantic classes: Surgical Tape (ST), Eye Retractors (ER), Instruments (Instru), Tissue Forceps (TF), I/A Handpiece (I/A H), Bonn Forceps (BF), Cap. Forceps (CF) in CaDIS. More results can be found in the supplementary video.} 
	\label{fig:results}
	\vspace{-1mm}
\end{figure*}

\begin{figure*}[!ht]
	\centering
	\includegraphics[width=0.98\textwidth]{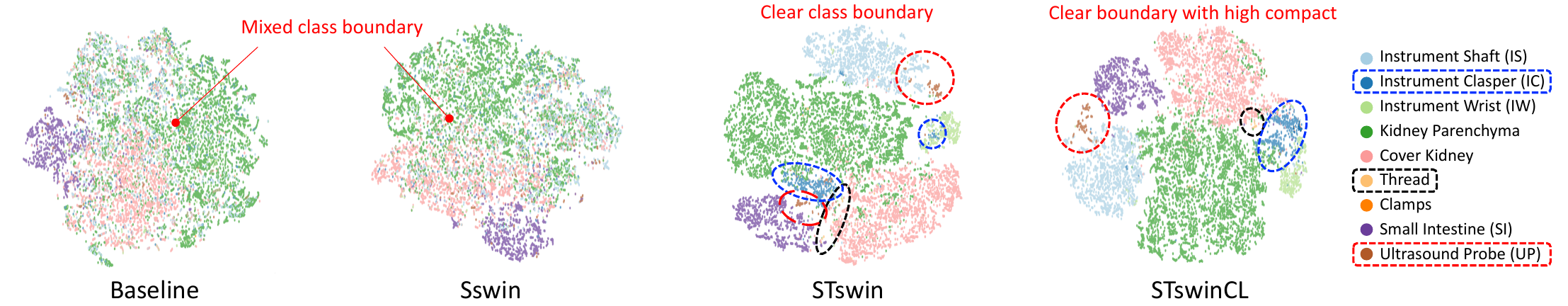}
	\vspace{-3mm}
	\caption{\textbf{Visualization of embedding spaces} learned under four settings on EndoVis18 dataset. Embeddings are colored according to class labels.}
	\label{fig:tsne}
	\vspace{-2mm}
\end{figure*}

\subsubsection{Qualitative comparison} 

In Fig.~\ref{fig:results}, we show the visual results of four settings on (a) EndoVis18 and (b-d) three tasks of CaDIS. 
Although not all the classes can be illustrated due to space limitation, we try to include more classes by selecting several typical frames.
We see that our STswinCL can predict accurate segmentations with complete masks on the class with high intra-class variance (cf. kidney in the first two rows of (a), Instru in (b), I/A H in (c))
It can also tackle the challenge of low inter-class variance, differentiating the classes with high similarity (cf. kidney and its cover in the 2nd row of EndoVis18, skin and ST in CaDIS).
STswinCL can also largely suppress the irrelevant and incorrect regions (cf. the 3rd row in (a)).
Additionally, all configurations can produce satisfactory results on the major classes. 
By gradually integrating two relation cues, STswinCL can achieve complete segmentations for the rare classes that seldom occur (cf. UP in the 3rd row of (a), TF in (c), BF in (d)).
STswinCL can also precisely identify the objects with small size, such as thread in the 1st row of (a), ER in (b) and CF in (d).
In Fig.~\ref{fig:tsne}, we further show t-SNE visualization of encoded embedding space learned from four settings on EndoVis18.
We ignore the background class for visualization, as it is officially excluded when evaluating the performance. In addition, the other two classes are unshown in this figure, including the suction-instrument class, which only exists in the training dataset; the needle, which only appears in few frames of validation dataset with extremely tiny size, therefore this class is highly difficult to be sampled for visualization.
We see that Sswin shows a better feature cluster on majority classes than Baseline, however, still suffers from the mixed class boundary.
STswin can facilitate feature discrimination with clearer boundary.
STswinCL further compacts each class and forms a well-structured space. Notably, some rare classes can also be attracted into their individual clusters (see black and red dashed circles).

\vspace{-2mm}
\subsection{Different Lengths of Supporting Frames}

Temporal context is a key factor in our joint space-time Transformer and we analyze the impact of different lengths of supporting frames which form the input $\bm{x}$ into the model. Specifically, we increase the number of frames with range $N$ $ \in [1,5]$.
The model is equivalent to Sswin when $N$ $=1$.
Inter-video contrast is not incorporated in these settings for a clear and direct comparison.
Experimental results on EndoVis18 dataset are reported in Table~\ref{tab:ab_length}.

We observe that compared with the single frame, improvements can be consistently gained when leveraging temporal support with various lengths.
The performance gradually improves from 61.7\% to 62.3\% at mIoU and from 70.4\% to 70.8\% at Dice, when $N$ increases from 1 to 4.
Nevertheless, we notice that 
overall result slightly decreases about 0.2\% at both mIoU and Dice, when length raising to 5.
The underlying reason may be that the distant frames present different appearances due to large surgical motion, bringing irrelevant noise disturbing accurate segmentation.
We therefore leverage 4-frame temporal information in our model learning.

We perform the statistical analysis by computing p-values using wilcoxon signed-rank test. Numbers are given in Table~\ref{tab:pvalue_length} when comparing all the other ablation settings to ours.
It is observed that we get $p < 0.05$ in mIoU for all settings, indicating the improvement significance about input length.

\begin{table}[!t]
\caption{Results on EndoVis18 with input clip in different lengths. }
\vspace{-2mm}
\label{tab:ab_length}
\centering
\resizebox{0.47\textwidth}{!}{
\begin{threeparttable}
\begin{tabular}{c|cccccc}
\hline
  \multirow{2}{*}{Frame} &
  \multirow{2}{*}{mIoU(\%)} & 
  \multicolumn{4}{c}{Sequence (mIoU(\%))} &
    \multirow{2}{*}{Dice(\%)} \\ \cline{3-6} 
                          &   & Seq1    & Seq2    & Seq3   & Seq4  &    \\ \hline
				1            &  61.7 & 63.8 & 59.7 & 83.2 & \textbf{40.0} & 70.4 \\
				2           & 62.0 &  65.0 & 61.0   &83.6  &38.2 & 70.7 \\
				3            & 62.1 & 63.8 & \textbf{61.3} & 84.1 & 38.8 & 70.7  \\
				\textbf{4}             & \textbf{62.3} &  \textbf{65.6} & 60.9 &  \textbf{84.3} & 38.5 &  \textbf{70.8} \\
				5           & 62.1  & 65.5  & 60.6  & 83.9 & 38.3   &  70.6 \\ \hline

\end{tabular}
\end{threeparttable}
}
\vspace{-3mm}
\end{table}

\begin{table}[t]
	\centering
	\caption{\textcolor{black}{P-values of ablations on input length (frames) for ours.}}
	\vspace{-1mm}
	\label{tab:pvalue_length}
		\resizebox{0.32\textwidth}{!}{
			\begin{tabular}{c|cccc}
				\hline
				Metrics      & 1    & 2     & 3 & 5 \\
				\hline
				mIoU    & 3e-5  & 0.002  & 0.011  & 0.033   \\
				Dice    & 5e-4  & 0.030  & 0.203  & 0.175   \\
				\hline
			\end{tabular}}
\end{table}

\vspace{-2mm}
\subsection{Detailed Analysis of Pair Construction}
The pair construction is a key factor in our inter-video contrastive learning, where the numbers of positive and negative pairs directly influence the scope of embedding space shaped in each training iteration.
Regarding the current frame as query, we study its effect on the segmentation results by varying the numbers of key frames, including the adjacent frames (mainly contribute to positives (Pos.)), and the frames from other videos (mainly contribute to negatives (Neg.)). We finally obtain six configurations.
For instance, $(0,0)$ suggests that contrast is solely performed within the current frame; (1,1) represents one adjacent frame and one inter-video frame are included; (-,-) is pure STswin without contrast.

Experimental results on the EndoVis18 dataset are presented in Table~\ref{tab:ab_pair}.
We see that solely relying on the current frame for contrast decreases results compared with pure STswin model, due to the limited contrast embedding space.
Even though only borrowing the adjacent frames to enlarge the regularized space, (1,0) achieves superior results with 1.7\% mIoU and 2.1\% Dice gain over (0,0).
While when only considering one frame from other videos, (1,1) slightly decreases results to 62.5\% mIoU.
Segmentation performances consistently improve from 62.5\% to 63.6\% by introducing more negatives from 1 to 3 frames, demonstrating that a richer set of negatives can benefit contrast process to learn a better representation.
However, result degradation is shown with the configurations of (1,4) and (2,3).
The underlying reason may be that imbalanced pos-neg pairs (negative bias in former and positive bias in latter) lead to difficulty in effectively contrastive training.
Note that another stream of contrastive learning methods, such as BYOL~\cite{grill2020bootstrap}, do not encounter the pair imbalance problem. As these methods do not reply on negative pairs by introducing an additional predictor in the regular branch.
They are promising especially when there exists high uncertainty in the negative pair construction.
In our work, negative pairs can be precisely formed under the guidance of ground truths. Therefore, we design our model to leverage the information from negative pairs via symmetric architecture following the MoCo stream~\cite{he2020momentum}. The balance between the pos-neg pairs is crucial for model training.
Generally, continuing to enlarge the positives and negatives with a balanced ratio shall further improve the results. However, we do need to consider a practical computation source for real-world usage, therefore, we choose (1,3) in our work.

We also conduct the statistical significance study about this key factor, with results shown in Table~\ref{tab:pvalue_pair}. We can see $p < 0.05$ in both evaluation metrics in all the settings, verifying the significance of pair construction for accurate scene segmentation.

\begin{table}[!t]
\caption{\textcolor{black}{Results on EndoVis18 with different Pos. and Neg. pairs. }}
\vspace{-2mm}
\label{tab:ab_pair}
\centering
\resizebox{0.49\textwidth}{!}{
\begin{threeparttable}
\begin{tabular}{cc|cccccc}
\hline
  \multirow{2}{*}{Pos.} &
  \multirow{2}{*}{Neg.} &
  \multirow{2}{*}{mIoU(\%)} & 
  \multicolumn{4}{c}{Sequence (mIoU(\%))} &
    \multirow{2}{*}{Dice(\%)} \\ \cline{4-7} 
                     &    &   & Seq1    & Seq2    & Seq3   & Seq4  &     \\ \hline
				- & -     & 62.3 &  65.6 & 60.9 &  \textbf{84.3} & 38.5 &  70.8 \\
				0 & 0      & 61.1  & 65.9  & 60.3 & 83.6  & 34.5  & 69.2  \\
				1 & 0 &  62.8  & 65.5  & 61.6 & 83.7  & 41.0 & 71.3  \\
				1 & 1       & 62.5  & 64.3  & 61.3 & 82.7  & \textbf{41.5} & 71.1  \\
				1 & 2            & 62.9  & 65.9  & 61.9 & 83.6  & 40.2 & 71.3  \\
				\textbf{1} & \textbf{3}   & \textbf{63.6} &  \textbf{67.0} & \textbf{63.4} &  83.7 & 40.3 &  \textbf{72.0}\\
				1 & 4       & 63.1  & \textbf{67.0} & 62.1 & 84.0 & 39.3 & 71.4   \\
				2 & 3            & 61.6  & 66.5  & 61.3 & 83.6  & 34.9 & 69.7  \\ \hline

\end{tabular}
\end{threeparttable}
}
\vspace{-3mm}
\end{table}

\begin{table}[t]
	\centering
	\caption{\textcolor{black}{P-values of all ablation settings \\on pair construction for our method.}}
	\vspace{-1mm}
	\label{tab:pvalue_pair}
		\resizebox{0.46\textwidth}{!}{
			\begin{tabular}{c|cccccc}
				\hline
				Metrics      & (0,0)     & (1,0)     & (1,1) & (1,2) & (1,4) & (2,3)\\
				\hline
				mIoU    & 3e-17  & 4e-4  & 3e-6  & 1e-5  & 0.007  & 9e-16  \\
				Dice    & 1e-9  & 7e-4  & 0.001  & 1e-4  & 0.003  & 3e-8  \\
				\hline
			\end{tabular}}
\end{table}

\vspace{-2mm}
\section{Discussion}

Automatic surgical scene segmentation plays a fundamental role in developing the next generation of CAI solutions.
While existing approaches rely on traditional feature aggregation modules, \eg, atrous convolution, ConvLSTM, they utilize only information within a local region.
We develop a STswin Transformer to efficiently relate more cues to augment the pixel representation, performing the local self-attention within a space-time window, and leveraging window shift to increase the aggregation range of both space and time dimensions. Computation cost decreases compared with the standard Transformer,
enabling each pixel to see a rich spatial (whole frame) and temporal (four adjacent frames in this work) view for the higher representation discrimination.

The standard segmentation models for surgical scenes learn to map pixels to an embedding space, and use CE loss to increase the category differentiation of pixel embedding towards the ground-truth class (pixel-to-ground-truth).
However, they still suffer from ambiguous decision boundaries in the embedding space, due to the high similarity of pixels from different classes, while limited similarity of pixels from the same class in surgical scene. 
We introduce a new pixel-to-pixel regularization, extending the relation modelling from current video to the whole dataset (cross videos), and obtain a better-structured space to tackle this primary challenge.
Integrating the proposed regularization to CE loss is crucial to complement each other.
One direct implementation is joint training the model towards the combination of these two objectives, however, it degrades performance into 70.6\% Dice and 62.0\% mIoU on EndoVis18. Our three-stage training strategy demonstrates the better way by employing the new regularization to pre-train the model, to yield a good initialization with high compact embedding space.
\textcolor{black}{The pre-train manner results in less influence of batch size in contrastive learning stage, on the final segmentation performance. Increasing batch size of current 4 to 6 and 8 does not bring the noticeable segmentation improvement.}

Our inter-video contrast can effectively tackle the class imbalance problem in surgical scene segmentation, with performance improvement in the rare classes. 
Pixel-to-pixel comparison avoids heavy dependence on the ground-truth labels, otherwise, the label bias naturally incurred in the imbalanced data shall drastically alter the decision boundary by the majority classes. Recent studies also point out that balancing classes in pair construction can benefit contrastive learning to generate a more robust feature space.
We utilized the label mask to suppress some pixels from majority classes when forming positive or negative pairs, also, tried to enlarge the weight of rare classes in similarity calculation.
However, the segmentation performances are similar to the current results without noticeable improvement ($\pm 0.2\%$ mIoU on EndoVis18).
\textcolor{black}{Therefore, we do not consider the pair re-sampling or re-weighting in the proposed pixel-to-pixel contrast.}
In future work, it is important to explore how to incorporate re-weighting and re-sampling strategies into final fine-tuning stage, to explicitly alleviate the label bias in affecting segmentation classifier by CE loss.

Our new training regime by concurrently considering both feature discrimination and space structure is flexible in network architecture and can be implemented on various segmentation models.
Our method on inter-video relation modelling could also be generalizable to many other scenarios, where temporal cue is the key factor, such as microscopy cell tracking, and cardiac motion estimation.
From experimental results, we see that \textcolor{black}{although our different configurations can perform consistently in mIoU and Dice, yield inconsistent efficacy across some surgical sequences. The inconsistency is also revealed in other methods (see results of different methods in Table~\ref{tab:result}.
The underlying reason may be the large variance of different surgical procedures given the complicated surgical scene.
The fairly limited dataset sizes may also lead to the unstable results. 
In future work, we plan to construct a larger-scale dataset for more comprehensive method validation.}

Our window shift scheme has dramatically relieved the computation burden when communicating richer cues. 
\textcolor{black}{Compared with DeepLabV3+ baseline, our Sswin only increases the inference time 0.005-0.015 second per frame with different datasets. 
While the standard self-attention mechanism without window shift increases the inference time  0.018-0.044 seconds per frame, and even performs worse on CaDIS.
The time growth purely from the time shift scheme is also marginal, with around 0.01s.
Moreover, our contrastive training does not introduce the extra cost during inference. 
To this end, the inference time of our complete model STSwinCL is around 0.191s and 0.073s per frame in EndoVis18 and CaDIS, respectively. Compared with the single-frame based model Baseline, the speed-accuracy tradeoff of our proposed method is promising, especially for CaDIS datasets (5.7\% mIoU yet 0.062s increase in Task III).}
\textcolor{black}{However, the general record frequency of surgical video is 25 fps and the proposed method currently cannot achieve the real-time inference.
A promising direction is further increasing inference efficiency for more practical usage in real-world surgery.}
In future efforts, we shall also explore knowledge distillation to learn a more lightweight student model, and active learning to reduce the number of involved frames by selecting more representative ones, for a cost-effective dense segmentation.

\section{Conclusion}
We proposed a novel STswinCL model for semantic segmentation of the full scene in surgical videos, which collaboratively models two types of relationships across different videos for accurate segmentation. Specifically, our designed Transformer with a joint space-time window shift scheme can efficiently relate the richer spatial and temporal content in intra video to each pixel, augmenting each embedding to higher-level discrimination.
STswinCL complements and mutually considers shaping the whole embedding space to a better structure by exploring inter-video pixel-to-pixel relation.
Promising experimental results on two typical surgical video datasets demonstrate the effectiveness of our method, presenting the superiority over prior state-of-the-art approaches.

\bibliographystyle{IEEEtran}
\small\bibliography{refs}

\end{document}